\documentclass[sigconf]{acmart}

\usepackage{footnote}
\makesavenoteenv{tabular}

\def\BibTeX{{\rm B\kern-.05em{\sc i\kern-.025em b}\kern-.08emT\kern-.1667em\lower.7ex\hbox{E}\kern-.125emX}}

\setcopyright{rightsretained}

\copyrightyear{2019}
\acmYear{2019}
\acmConference[MM '19]{Proceedings of the 27th ACM International Conference on Multimedia}{October 21--25, 2019}{Nice, France}
\acmBooktitle{Proceedings of the 27th ACM International Conference on Multimedia (MM '19), October 21--25, 2019, Nice, France}
\acmPrice{}
\acmDOI{10.1145/3343031.3350539}
\acmISBN{978-1-4503-6889-6/19/10}

\def\repourl{https://github.com/xdspacelab/openvslam}

\begin{document}

\fancyhead{}

\title{OpenVSLAM: A Versatile Visual SLAM Framework}

\author{Shinya Sumikura}
\affiliation{
  \institution{Nagoya University}
  \state{Aichi}
  \country{Japan}
}
\email{sumikura@ucl.nuee.nagoya-u.ac.jp}

\author{Mikiya Shibuya}
\affiliation{
  \institution{Tokyo Institute of Technology}
  \state{Tokyo}
  \country{Japan}
}
\email{shibuya.m.ab@m.titech.ac.jp}

\author{Ken Sakurada}
\affiliation{
  \institution{National Institute of Advanced Industrial Science and Technology}
  \state{Tokyo}
  \country{Japan}
}
\email{k.sakurada@aist.go.jp}

\begin{abstract}
In this paper, we introduce OpenVSLAM, a visual SLAM framework with high usability and extensibility.
Visual SLAM systems are essential for AR devices, autonomous control of robots and drones, etc.
However, conventional open-source visual SLAM frameworks are not appropriately designed as libraries called from third-party programs.
To overcome this situation, we have developed a novel visual SLAM framework.
This software is designed to be easily used and extended.
It incorporates several useful features and functions for research and development.
\end{abstract}

\copyrightyear{2019}
\acmYear{2019}
\acmConference[MM '19]{Proceedings of the 27th ACM International Conference on Multimedia}{October 21--25, 2019}{Nice, France}
\acmBooktitle{Proceedings of the 27th ACM International Conference on Multimedia (MM '19), October 21--25, 2019, Nice, France}
\acmPrice{}
\acmDOI{10.1145/3343031.3350539}
\acmISBN{978-1-4503-6889-6/19/10}

\ccsdesc[500]{Software and its engineering~Software libraries and repositories}
\ccsdesc[300]{Computing methodologies~Scene understanding}
\ccsdesc[300]{Computing methodologies~Vision for robotics}

\keywords{Visual SLAM; Visual Odometry; Scene Modeling; Scene Mapping; Localization; Open Source Software; Computer Vision}

\begin{teaserfigure}
  \centering
  \includegraphics[height=4.1cm]{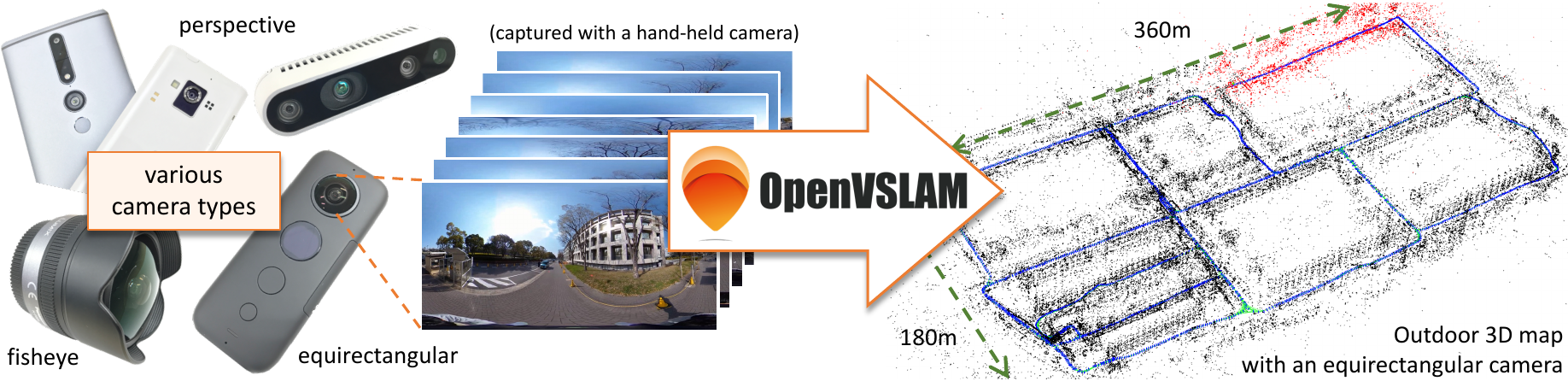} 
  \caption{One of the noteworthy features of OpenVSLAM: 3D scene mapping with various types of camera models.}
  \Description{}
  \vspace{0.26cm}
  \label{fig:teaser}
\end{teaserfigure}

\maketitle

\section{Introduction}

Simultaneous localization and mapping (SLAM) systems have experienced a notable and rapid progression through enthusiastic research and investigation conducted by researchers in the fields of computer vision and robotics.
In particular, ORB--SLAM~\cite{murartal2015orbslam,murartal2017orbslam2}, LSD--SLAM~\cite{engel2014lsdslam}, and DSO~\cite{engel2018direct} constitute major approaches regarded as de facto standards of visual SLAM, which performs SLAM processing using imagery.
These approaches have achieved state-of-the-art performance as visual SLAM.
In addition, researchers can reproduce the behavior of these systems on their computers because their source code is open to the public.
However, they are not appropriately designed in terms of usability and extensibility as visual SLAM libraries.
Thus, researchers and engineers have to make a great effort to apply those SLAM systems to their applications.
In other words, it is inconvenient to use existing open-source software (OSS) for visual SLAM as the basis of applications derived from 3D modeling and mapping techniques, such as autonomous control of robots and unmanned aerial vehicles (UAVs), and augmented reality (AR) on mobile devices.
Therefore, it is definitely valuable to provide an open-source visual SLAM framework that is easy to use and to extend by users of visual SLAM.

In this paper, we present OpenVSLAM, a monocular, stereo, and RGBD visual SLAM system that comprises well-known SLAM approaches, encapsulating them in several separated components with clear application programming interfaces (APIs).
We also provide extensive documentation for it, including sample code snippets.
The main contributions of OpenVSLAM are
\begin{itemize}
  \item It is compatible with various types of camera models and can be customized for optional camera models.
  \item Created maps can be stored and loaded, then OpenVSLAM can localize new images using prebuilt maps.
  \item A cross-platform viewer running on web browsers is provided for convenience of users.
\end{itemize}
\begin{savenotes}
    \begin{table*}[t]
      \centering
      \caption{Comparison of several open-source visual SLAM frameworks.}
      \label{table:comparison_slams}
      \vspace{-8pt}
      \begin{tabular}{c|cccccc}
        \hline
        & ORB--SLAM2~\cite{murartal2017orbslam2} & LSD--SLAM~\cite{engel2014lsdslam} & DSO~\cite{engel2018direct} & ProSLAM~\cite{schlegel2018proslam} & UcoSLAM~\cite{rafael2019ucoslam} & \textbf{OpenVSLAM} (ours) \\
        \hline
        OSS license & GPLv3 & GPLv3 & GPLv3 & 3-clause BSD & GPLv3 &  - \footnote{See \url{\repourl} about the license.} \\
        \hline
        SLAM method & indirect & direct & direct & indirect & indirect + marker & indirect \\
        \hline
        camera model & perspective & perspective & perspective & perspective & perspective &
        \begin{tabular}{c}
          perspective, fisheye, \\ equirectangular
        \end{tabular} \\
        \hline
        setup &
        \begin{tabular}{c}
          monocular, \\ stereo, RGBD
        \end{tabular} &
        monocular & monocular & stereo, RGBD &
        \begin{tabular}{c}
          monocular, \\ stereo, RGBD
        \end{tabular} &
        \begin{tabular}{c}
          monocular, \\ stereo, RGBD
        \end{tabular} \\
        \hline
        map store/load &   &   &   &   & \checkmark & \checkmark \\
        \hline
        customizability &   &   &   & \checkmark &   & \checkmark \\
        \hline
      \end{tabular}
    \end{table*}
\end{savenotes}
One of the noteworthy features of OpenVSLAM is that the system can deal with various types of camera models, such as perspective, fisheye, and equirectangular, as shown in Figure\,\ref{fig:teaser}.
AR on mobile devices such as smartphones needs a SLAM system with a regular perspective camera.
Meanwhile, fisheye cameras are often mounted on UAVs and robots for visual SLAM and scene understanding because they have a wider field of view (FoV) than perspective cameras.
OpenVSLAM can be used with almost the same implementation between perspective and fisheye camera models.
In addition, it is a significant contribution that equirectangular images can constitute inputs to our SLAM system.
By using cameras that can capture omnidirectional imagery, the tracking performance of SLAM can be improved.
Our efforts to make use of equirectangular images for visual SLAM enable tracking and mapping not to depend on the direction of a camera.
Furthermore, OpenVSLAM provides interfaces that can be employed for applications and researches that use visual SLAM.
For example, our SLAM system incorporates interfaces to store and load a map database and a localization function based on a prebuilt map.

We contribute to the community of computer vision and robotics by providing this SLAM framework, as shown in Table\,\ref{table:comparison_slams}, so that researchers can jointly contribute to its development.

\section{Related Work}
\label{sec:related_work}

\subsection{OSS for Scene Modeling}

In this section, mapping and localization techniques whose programs are released as OSS are briefly described.
Such techniques are essential in a wide variety of application scenarios for autonomous control of UAVs and robots, AR on mobile devices, etc.
Some OSS packages for those tasks using images have been open to the public.

Structure from motion (SfM) and visual SLAM are often employed as scene modeling techniques based on imagery.
Regarding SfM, it is usually assumed that the entire image set is prepared in advance.
Then the algorithm performs 3D reconstruction via batch processing.
Concerning visual SLAM, 3D reconstruction is processed in real-time.
Therefore, it assumes that images are sequentially input.
OpenMVG~\cite{moulon2019openmvg}, Theia~\cite{sweeney2015theia}, OpenSfM~\cite{mapillary2019opensfm}, and COLMAP~\cite{schoenberger2016sfm} are well-known OSS packages for SfM.
Some SfM frameworks~\cite{moulon2019openmvg,mapillary2019opensfm} are capable of dealing with fisheye and equirectangular imagery.
The compatibility with such images has improved the performance and usability of SfM packages as 3D modeling frameworks.
Meanwhile, researchers often use visual SLAM, such as ORB--SLAM~\cite{murartal2015orbslam,murartal2017orbslam2}, LSD--SLAM~\cite{engel2014lsdslam}, and DSO~\cite{engel2018direct}, for real-time 3D mapping.
Unlike some SfM frameworks, most of the visual SLAM software programs can only handle perspective imagery.
In our case, inspired by the aforementioned SfM frameworks, we do provide a novel visual SLAM framework compatible with various types of camera models.
We thus aim at improving usability and extensibility of visual SLAM for 3D mapping and localization.

\subsection{Visual SLAM}

Some visual SLAM programs are introduced and some of their features are explained in this section.
Table\,\ref{table:comparison_slams} compares characteristics of well-known visual SLAM frameworks with our OpenVSLAM.

ORB--SLAM~\cite{murartal2015orbslam,murartal2017orbslam2} is a kind of indirect SLAM that carries out visual SLAM processing using local feature matching among frames at different time instants.
In this approach, the FAST algorithm~\cite{rosten2006machine,rosten2010faster} is used for keypoint detection.
The binary vector~\cite{rublee2011orb} is then used for its descriptor.
Quick methods that can extract keypoints and match feature vectors enable visual SLAM algorithms to be processed in real-time.
Similar approaches are employed in ProSLAM \cite{schlegel2018proslam}, which is the simple visual SLAM framework for perspective stereo and RGBD camera systems.
UcoSLAM~\cite{rafael2019ucoslam} adopts an algorithm that combines artificial landmarks, such as squared fiducial markers, and binary descriptor used by ORB--SLAM and ProSLAM.
Meanwhile, LSD--SLAM~\cite{engel2014lsdslam} and DSO~\cite{engel2018direct} are two different approaches of direct SLAM, which realizes visual SLAM processing directly exploiting brightness information of each pixel in images.
It should be noted that the direct method does not have to explicitly extract any keypoints from images.
Unlike the indirect method, the direct method can be correctly operated in more texture-less environments because it utilizes whole information from images.
However, the direct method presents more susceptibility to changes in lighting conditions.
Additionally, it has been reported that the direct method achieves lower performance than the indirect one when using rolling shutter cameras~\cite{engel2014lsdslam,engel2018direct}.
Given that image sensors in smartphones and consumer cameras are rolling shutter, OpenVSLAM adopts the indirect method for visual SLAM.

Most of the visual SLAM frameworks cannot store and load map databases, as highlighted in Table\,\ref{table:comparison_slams}.
Localization based on a prebuilt map is important in practical terms for a lot of application.
Accordingly, it is clear that the ability to store and load created maps improves the usability and extensibility of a visual SLAM framework.
Therefore, functions for I/O of map databases are implemented in OpenVSLAM.

\section{Implementation}
\label{sec:implementation}

\begin{figure}[t]
  \centering
  \includegraphics[width=\linewidth]{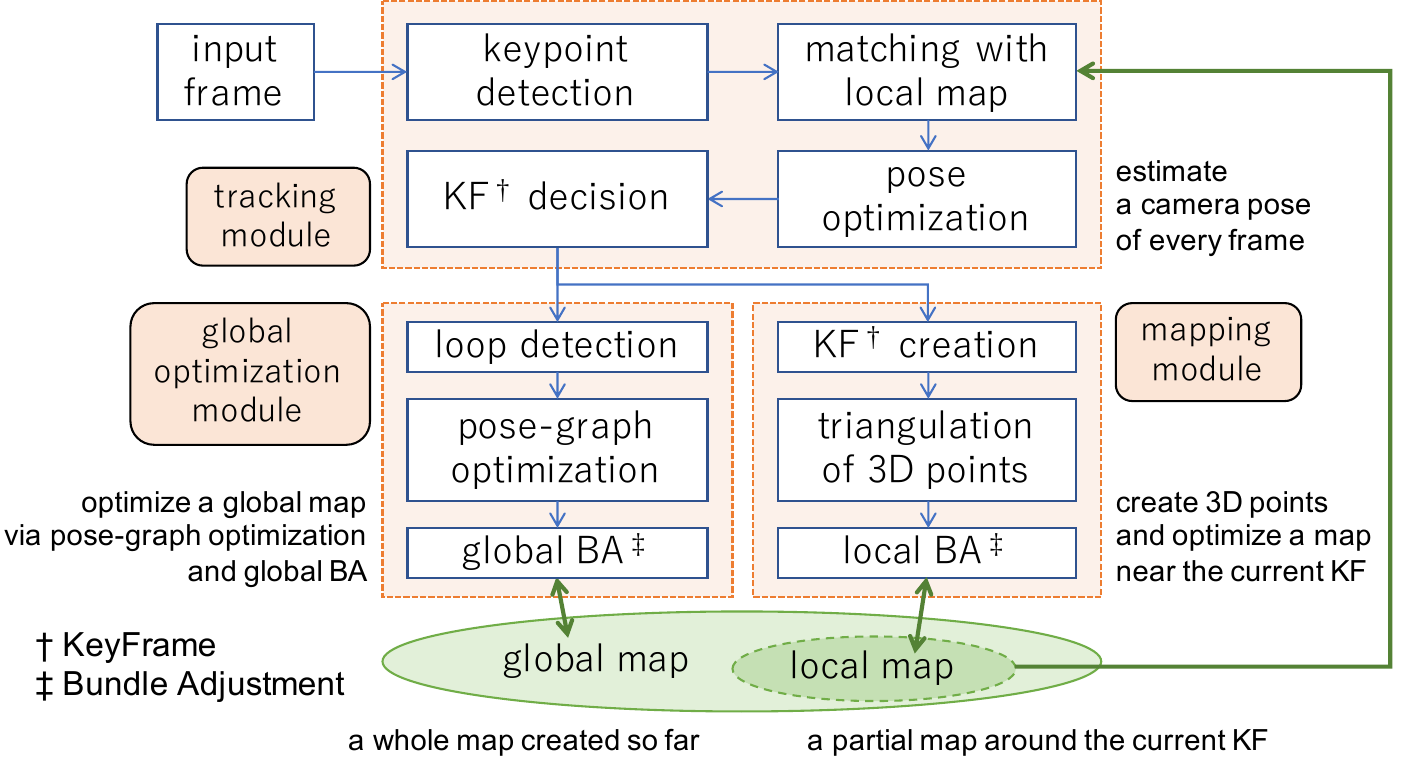}

  \vspace{-5pt}

  \caption{Main modules of OpenVSLAM: tracking, mapping, and global optimization modules.}
  \Description{}
  \label{fig:software_structure}
\end{figure}

OpenVSLAM is mainly implemented with C++.
It includes well-known libraries, such as Eigen\footnote{C++ template library for linear algebra: \url{http://eigen.tuxfamily.org/}} for matrix calculation, OpenCV\footnote{\label{fn:opencv}Open Source Computer Vision Library: \url{http://opencv.org/}} for I/O operation of images and feature extraction, and g2o~\cite{kummerle2011g2o} for map optimization.
In addition, extensive documentation including sample code snippets is provided.
Users can employ these snippets for their programs.

\subsection{SLAM Algorithm}

In this section, we present a brief outline of the SLAM algorithm adopted by OpenVSLAM and its module structure.
As in ORB--SLAM~\cite{murartal2015orbslam,murartal2017orbslam2} and ProSLAM~\cite{schlegel2018proslam}, the graph-based SLAM algorithm~\cite{grisetti2010tutorial} with the indirect method is used in OpenVSLAM.
It adopts ORB~\cite{rublee2011orb} as a feature extractor.
The module structure of OpenVSLAM is carefully designed for the customizability.

The software of OpenVSLAM is roughly divided into three modules, as shown in Figure\,\ref{fig:software_structure}: tracking, mapping, and global optimization modules.
The tracking module estimates a camera pose for every frame that is sequentially inputted to OpenVSLAM via keypoint matching and pose optimization.
This module also decides whether to insert a new keyframe (KF) or not.
When a frame is regarded as appropriate for a new KF, it is sent to the mapping and the global optimization modules.
In the mapping module, new 3D points are triangulated using the inserted KFs; that is, the map is created and extended.
Additionally, the windowed map optimization, called local bundle adjustment (BA), is performed in this module.
Loop detection, pose-graph optimization, and global BA are carried out in the global optimization module.
Trajectory drift, which often becomes a problem in SLAM, is resolved via pose-graph optimization implemented with g2o~\cite{kummerle2011g2o}.
Scale drift is also canceled in this way, especially for monocular camera models.

\subsection{Camera Models}

OpenVSLAM can accept images captured with perspective, fisheye, and equirectangular cameras.
In regard to perspective and fisheye camera models, the framework is compatible not only with monocular but also with stereo and RGBD setups.
Additionally, users can easily add new camera models (e.g., dual fisheye and catadioptric) by implementing new camera model classes derived from a base class \texttt{camera::base}.
This is a great advantage compared to other SLAM frameworks because new camera models can be implemented easily.

It is a noteworthy point that OpenVSLAM can perform SLAM with an equirectangular camera.
Equirectangular cameras, such as RICOH THETA series, insta360 series, and Ladybug series, have been recently used to capture omnidirectional images and videos.
In regard to visual SLAM, being compatible with equirectangular cameras implies a significant benefit for tracking and mapping because they have omnidirectional view, unlike perspective ones.
To the best of our knowledge, this is the first open-source visual SLAM framework that can accept equirectangular imagery.

\subsection{Map I/O and Localization}

As opposed to most of the visual SLAM frameworks, OpenVSLAM has functions to store and load map information, as shown in Table\,\ref{table:comparison_slams}.
In addition, users can localize new frames based on a prebuilt map.
The map database is stored in MessagePack format, hence the map information can be reused for any third-party applications in addition to OpenVSLAM.

\section{Quantitative Evaluation}
\label{sec:quantitative_evaluation}

In this section, tracking accuracy of OpenVSLAM is evaluated using EuRoC MAV dataset~\cite{burri2016euroc} and KITTI Odomery dataset~\cite{geiger2012arewe}, both of which have ground-truth trajectories.
ORB--SLAM2~\cite{murartal2017orbslam2}, the typical indirect SLAM, is selected for comparison.
In addition to tracking accuracy, tracking times are also compared.

Absolute trajectory error (ATE)~\cite{sturm2012benchmark} is used for evaluation of estimated trajectories.
To align an estimated trajectory and the corresponding ground-truth, transformation parameters between the two trajectories are estimated using Umeyama's method~\cite{umemiya1991least}.
$\mathrm{Sim(3)}$ transformation is estimated for monocular sequences because tracked trajectories are up-to-scale.
On the contrary, $\mathrm{SE(3)}$ transformation is used for stereo sequences.
The laptop computer used for the evaluations equips a Core i7-7820HK CPU (2.90GHz, 4C8T) and 32GB RAM.

\subsection{EuRoC MAV Dataset}
\label{subsec:euroc_mav_dataset_eval}

\begin{figure}[t]
  \centering
  \includegraphics[height=3.0cm]{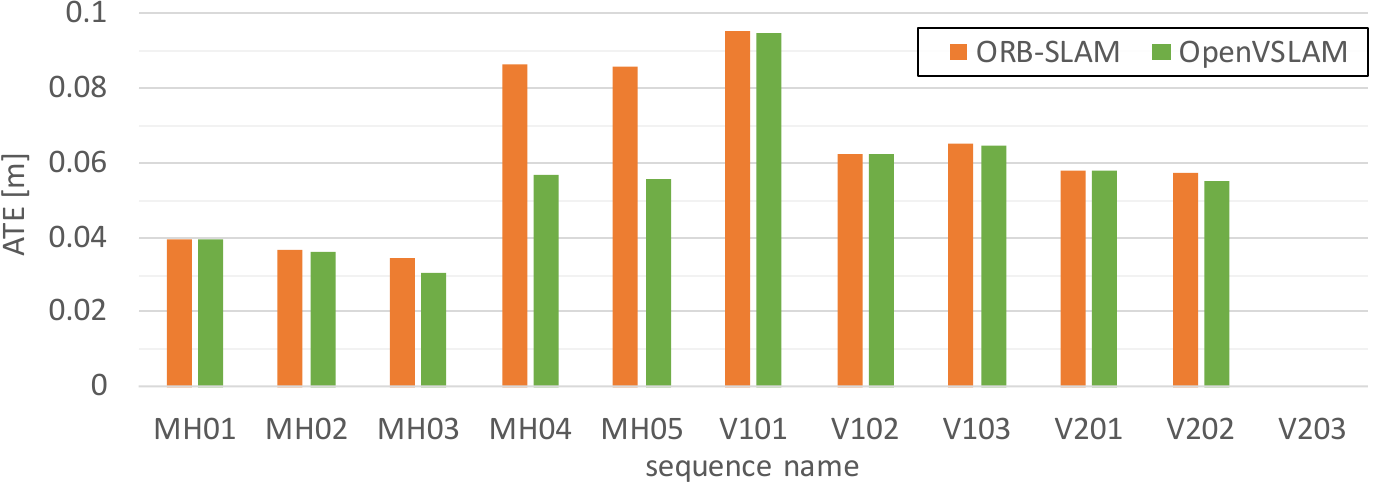} 

  \vspace{-5pt}

  \caption{Absolute trajectory errors on the 11 sequences in EuRoC MAV dataset (monocular). Lower is better.}
  \Description{}
  \label{fig:euroc_mono_ate}
\end{figure}

Figure\,\ref{fig:euroc_mono_ate} shows ATEs on the 11 sequences of EuRoC MAV dataset.
From the graph, it is found that OpenVSLAM is comparable to ORB--SLAM with respect to tracking accuracy for UAV-mounted cameras.
Concerning the sequences including dark scenes (\texttt{MH\_04} and \texttt{MH\_05}), the trajectories estimated with OpenVSLAM are more accurate than that with ORB--SLAM.
This is mainly because frame tracking method based on robust matching is additionally implemented in OpenVSLAM.

\begin{figure}[t]
  \centering

  \small
  \begin{tabular}{r|cc}
    \hline
    & ORB--SLAM & OpenVSLAM \\
    \hline
    mean [ms/frame] & 27.96 & \textbf{23.84} \\
    median [ms/frame] & 24.97 & \textbf{23.38} \\
    \hline
  \end{tabular}

  \vspace{2pt}

  \includegraphics[height=2.9cm]{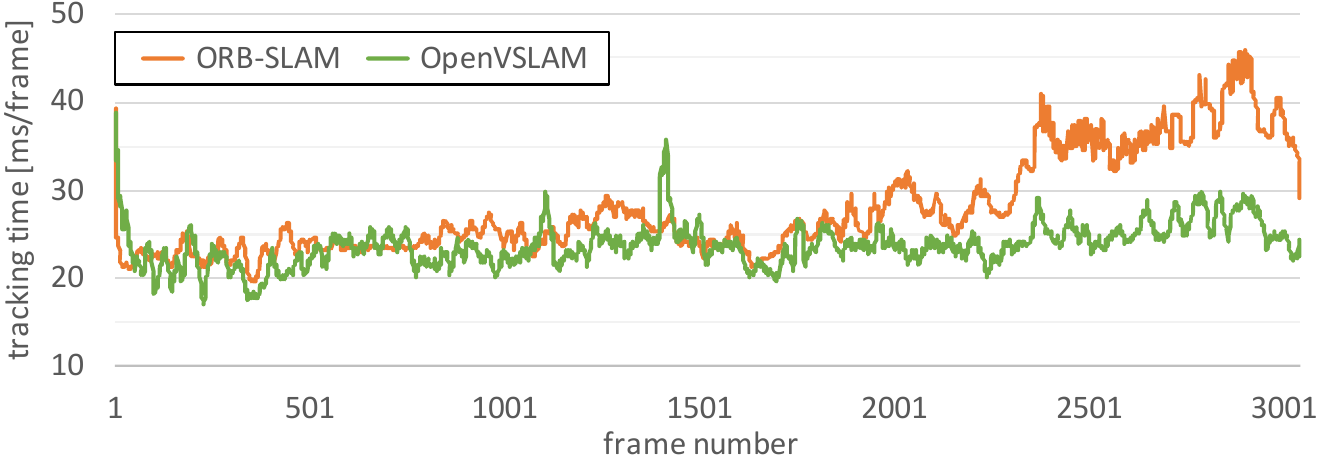}

  \vspace{-5pt}

  \caption{Tracking times on the \texttt{MH\_02} sequence of EuRoC MAV dataset (monocular). The table shows mean and median tracking times on each of the two frameworks. The graph shows the change in tracking times. Lower is better.}
  \Description{}
  \label{fig:euroc_tracking_time}
\end{figure}

Subsequently, tracking times measured using the \texttt{MH\_02} sequence of EuRoC MAV dataset are shown in Figure\,\ref{fig:euroc_tracking_time}.
Mean and median tracking times are presented in the table as well.
From the table, it is found that OpenVSLAM consumes less tracking time than ORB--SLAM.
This is mainly because the implementation of ORB extraction in OpenVSLAM is more optimized than that in ORB--SLAM.
In addition, it should be noted that OpenVSLAM requires less tracking time than ORB--SLAM in later parts of the sequence as shown in the graph.
This is because OpenVSLAM efficiently prevents a local map from being enlarged in the tracking module when a global map is expanded.

\subsection{KITTI Odometry Dataset}

\begin{figure}[t]
  \centering
  \includegraphics[height=3.0cm]{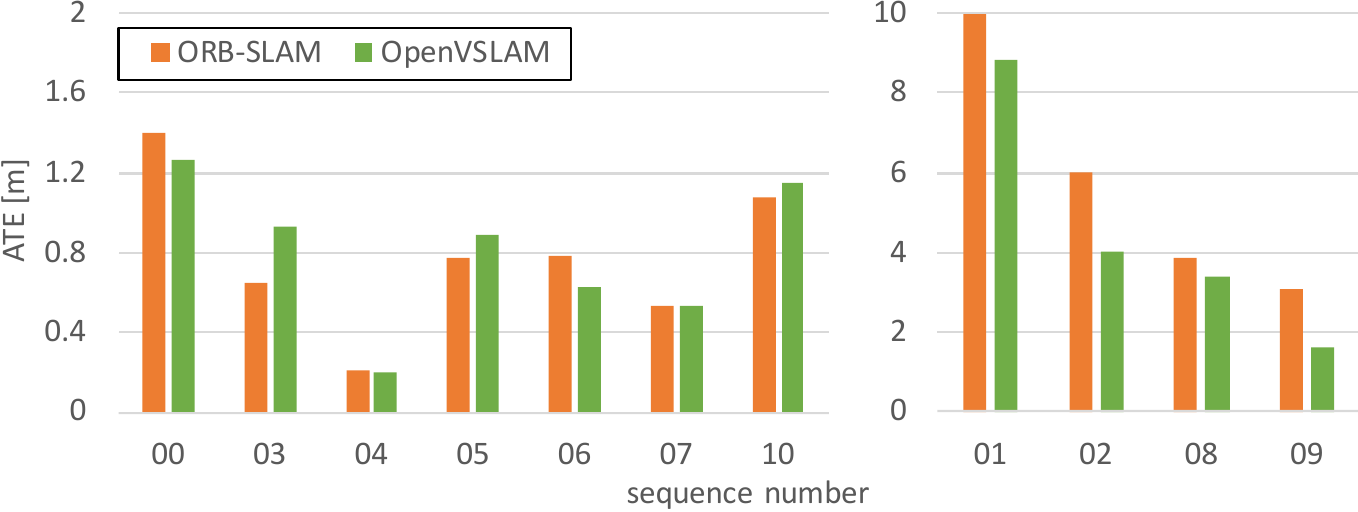} 

  \vspace{-5pt}

  \caption{Absolute trajectory errors on the 11 sequences in KITTI Odometry dataset (stereo). Lower is better.}
  \Description{}
  \label{fig:kitti_stereo_ate}
\end{figure}

Figure\,\ref{fig:kitti_stereo_ate} shows ATEs on the 11 sequences of KITTI Odometry dataset.
From the graph, it is found that OpenVSLAM has comparable performance to ORB--SLAM with respect to tracking accuracy for car-mounted cameras.

\begin{figure}[t]
  \centering

  \small
  \begin{tabular}{r|cc}
    \hline
    & ORB--SLAM & OpenVSLAM \\
    \hline
    mean [ms/frame] & 68.78 & \textbf{56.32} \\
    median [ms/frame] & 66.78 & \textbf{54.45} \\
    \hline
  \end{tabular}

  \vspace{2pt}

  \includegraphics[height=2.9cm]{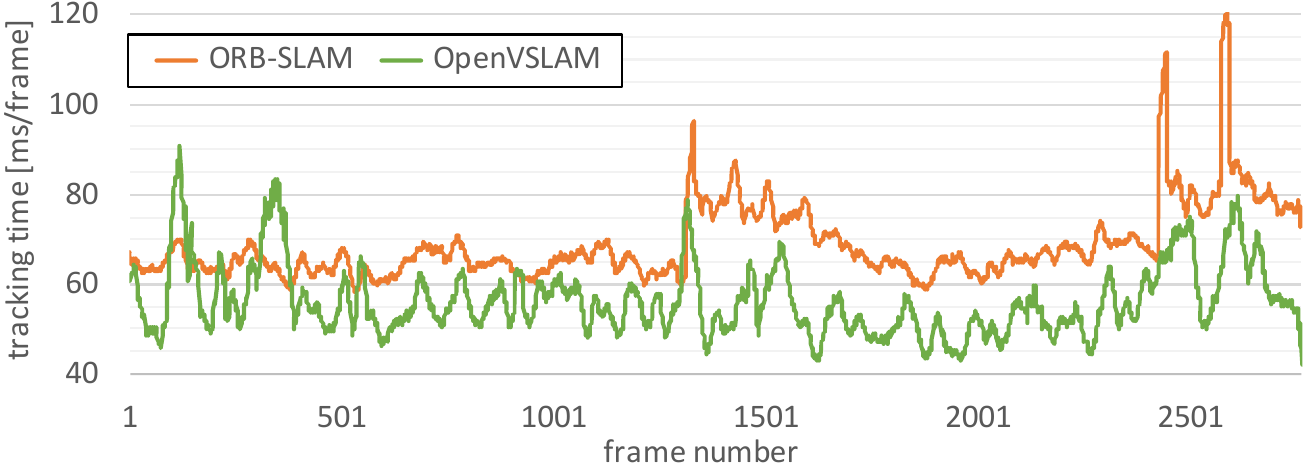}

  \vspace{-5pt}

  \caption{Tracking times on the sequence number \texttt{05} of KITTI Odometry dataset (stereo). The table shows mean and median tracking times on each of the two frameworks. The graph shows the change in tracking times. Lower is better.}
  \Description{}
  \label{fig:kitti_tracking_time}
\end{figure}

Subsequently, tracking times measured using the sequence number \texttt{05} of KITTI Odometry dataset are shown in Figure\,\ref{fig:kitti_tracking_time}.
Mean and median tracking times are also presented in the table.
OpenVSLAM consumes less tracking time than ORB--SLAM for the same reason described in Section\,\ref{subsec:euroc_mav_dataset_eval}.
The difference in the tracking times shown in Figure\,\ref{fig:kitti_tracking_time} is bigger than that in Figure\,\ref{fig:euroc_tracking_time} because 1) image size of KITTI Odometry dataset is larger than that of EuRoC MAV dataset, and 2) stereo-matching implementation in OpenVSLAM is also more optimized than that in ORB--SLAM.

\section{Qualitative Results}
\label{sec:qulitative_evaluation}

\subsection{Fisheye Camera}

In this section, experimental results of visual SLAM with a fisheye camera are presented both outdoors and indoors.
A LUMIX DMC-GX8 which equips an 8mm fisheye lens (Panasonic Corp.) is used for capturing image sequences.
The FPS of the camera is $30.0$.

\begin{figure}[t]
  \centering
  \includegraphics[width=\linewidth]{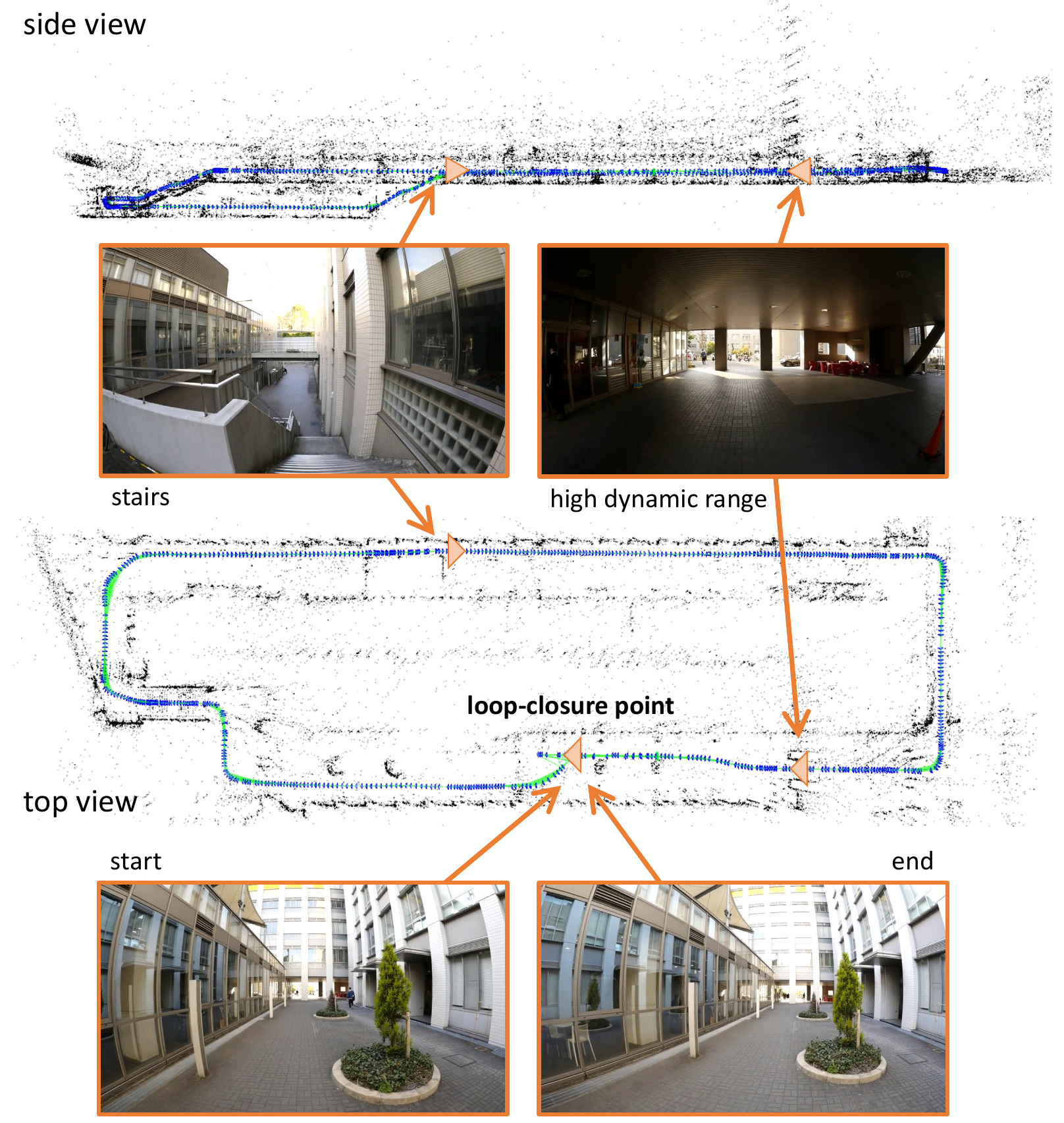} 

  \vspace{-5pt}

  \caption{Mapping result using the outdoor fisheye video. The top image depicts side-view of the 3D map, while the center image is top-view. The difference in elevation can be observed from the side-view.}
  \Description{}
  \label{fig:fisheye_slam_outdoor}
\end{figure}

The 3D map shown in Figure\,\ref{fig:fisheye_slam_outdoor} is created with OpenVSLAM using a fisheye video captured outdoor.
The number of frames is about $6400$.
The difference in elevation is observed from the side-view of the 3D map.
Also, it should be noted that camera pose tracking succeeded even in high dynamic range scenes.

\begin{figure}[t]
  \centering
  \includegraphics[width=7.0cm]{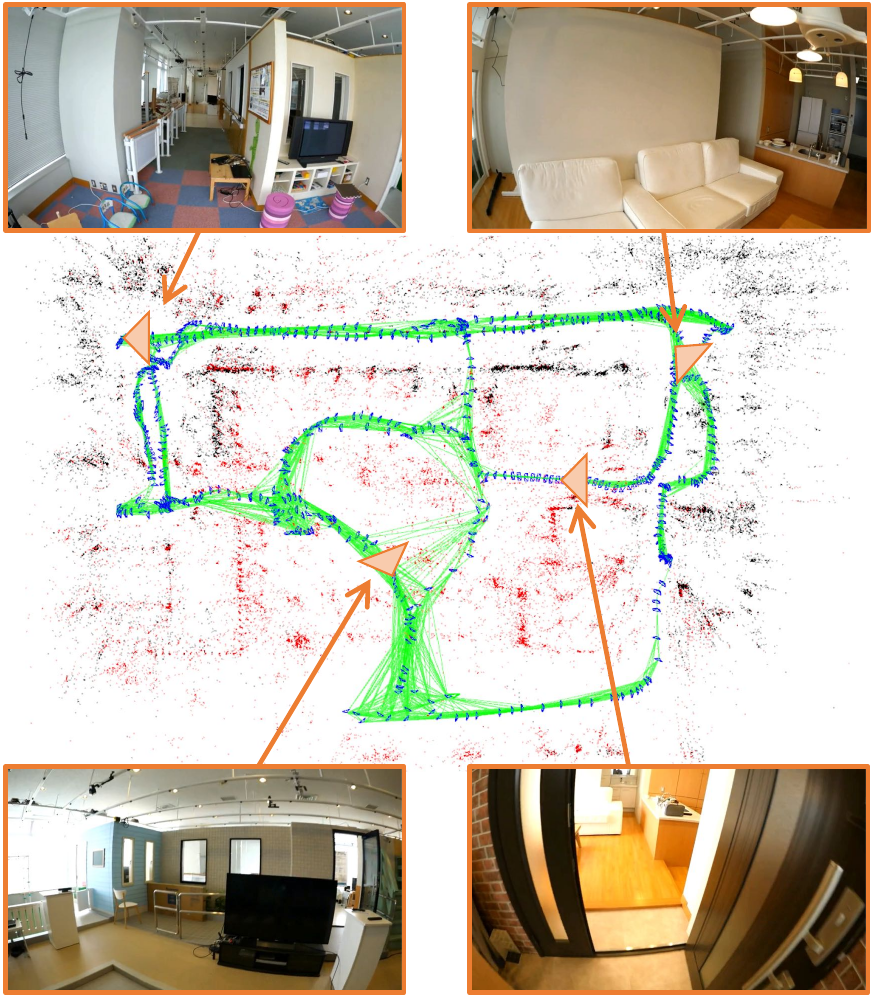} 

  \vspace{-5pt}

  \caption{Mapping result using the indoor fisheye video. The center image depicts the top-view of the 3D map. It is found that the shape of the room is reconstructed well.}
  \Description{}
  \label{fig:fisheye_slam_indoor}
\end{figure}

Figure\,\ref{fig:fisheye_slam_indoor} presents the 3D map built from an indoor fisheye video.
The number of frames is about $6700$.
It is found that the shape of the room is reconstructed clearly.
In addition, tracking succeeded even in areas that have less common views (e.g., the right-bottom image of Figure\,\ref{fig:fisheye_slam_indoor}).

These results allow us to conclude that visual SLAM with fisheye cameras is correctly performed both outdoors and indoors.

\subsection{Equirectangular Camera}

In this section, experimental results of visual SLAM with a THETA V (RICOH Co., Ltd.), a consumer equirectangular camera, are presented.

The 3D map shown in the right half of Figure\,\ref{fig:teaser} is created with OpenVSLAM using an equirectangular video captured outdoor.
The FPS is $10.0$ and the number of frames is $15000$.
It is found that tracking of camera movement, loop-closing, and global optimization work well even for the large-scale sequence.

\begin{figure}[t]
  \centering
  \includegraphics[width=7.0cm]{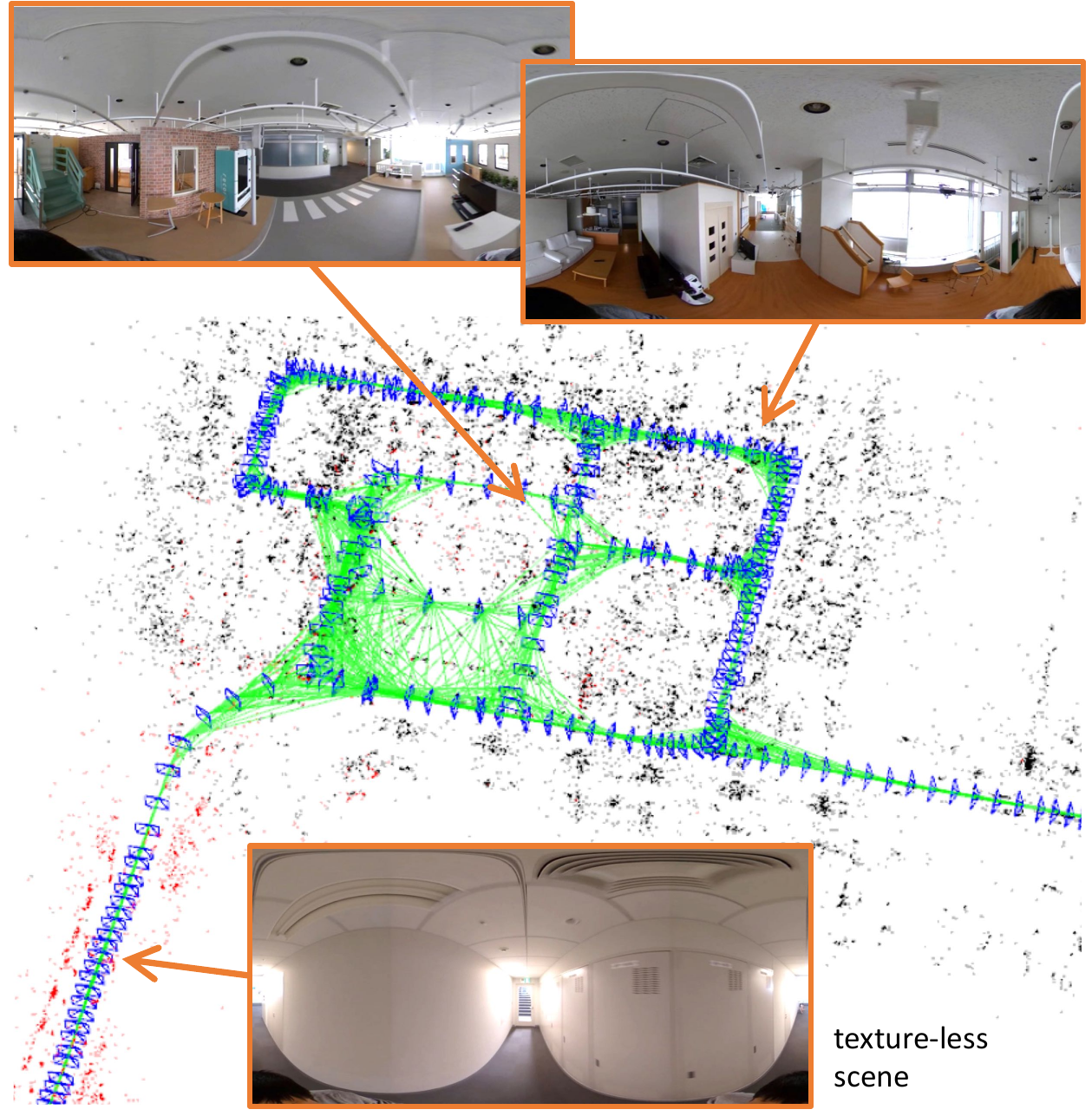} 

  \vspace{-5pt}

  \caption{Mapping result using the indoor equirectangular video. Tracking succeeds even in the texture-less areas.}
  \Description{}
  \label{fig:equirectangular_slam_indoor}
\end{figure}

Meanwhile, Figure\,\ref{fig:equirectangular_slam_indoor} presents the 3D map based on an indoor equirectangular video.
In this case, the FPS is $10.0$ and the number of frames is $1430$.
It should be noted that the camera poses are correctly tracked even in texture-less areas thanks to omnidirectional observation.

These results allow us to conclude that visual SLAM with equirectangular cameras is correctly performed both outdoors and indoors.

\vspace{6pt}

\section{Conclusion}
\label{sec:conclusion}

In this project, we have developed OpenVSLAM, a visual SLAM framework with high usability and extensibility.
The software is designed to be easily used for various application scenarios of visual SLAM.
It incorporates several useful functions for research and development.
In this paper, the quantitative performance is evaluated using the benchmarking dataset.
In addition, experimental results of visual SLAM with fisheye and equirectangular camera models are presented.
We will continuously maintain this framework for the further development of computer vision and robotics fields.

\begin{acks}
The authors would like to thank Mr. H. Ishikawa, Mr. M. Ichihara, Dr. M. Onishi, Dr. R. Nakamura, and Prof. N. Kawaguchi, for their support for this project.
\end{acks}

%
\bibliographystyle{acmref}
\bibliography{acmmm}


\begin{thebibliography}{19}


\ifx \showCODEN    \undefined \def \showCODEN     #1{\unskip}     \fi
\ifx \showDOI      \undefined \def \showDOI       #1{#1}\fi
\ifx \showISBNx    \undefined \def \showISBNx     #1{\unskip}     \fi
\ifx \showISBNxiii \undefined \def \showISBNxiii  #1{\unskip}     \fi
\ifx \showISSN     \undefined \def \showISSN      #1{\unskip}     \fi
\ifx \showLCCN     \undefined \def \showLCCN      #1{\unskip}     \fi
\ifx \shownote     \undefined \def \shownote      #1{#1}          \fi
\ifx \showarticletitle \undefined \def \showarticletitle #1{#1}   \fi
\ifx \showURL      \undefined \def \showURL       {\relax}        \fi
\providecommand\bibfield[2]{#2}
\providecommand\bibinfo[2]{#2}
\providecommand\natexlab[1]{#1}
\providecommand\showeprint[2][]{arXiv:#2}

\bibitem[\protect\citeauthoryear{AB}{AB}{2019}]%
        {mapillary2019opensfm}
\bibfield{author}{\bibinfo{person}{Mapillary AB}.}
  \bibinfo{year}{2019}\natexlab{}.
\newblock \bibinfo{title}{OpenSfM}.
\newblock \bibinfo{howpublished}{\url{https://github.com/mapillary/OpenSfM}}.
\newblock


\bibitem[\protect\citeauthoryear{Burri, Nikolic, Gohl, Schneider, Rehder,
  Omari, Achtelik, and Siegwart}{Burri et~al\mbox{.}}{2016}]%
        {burri2016euroc}
\bibfield{author}{\bibinfo{person}{Michael Burri}, \bibinfo{person}{Janosch
  Nikolic}, \bibinfo{person}{Pascal Gohl}, \bibinfo{person}{Thomas Schneider},
  \bibinfo{person}{Joern Rehder}, \bibinfo{person}{Sammy Omari},
  \bibinfo{person}{Markus~W Achtelik}, {and} \bibinfo{person}{Roland
  Siegwart}.} \bibinfo{year}{2016}\natexlab{}.
\newblock \showarticletitle{The EuRoC micro aerial vehicle datasets}.
\newblock \bibinfo{journal}{\emph{International Journal of Robotics Research
  (IJRR)}} \bibinfo{volume}{35}, \bibinfo{number}{10} (\bibinfo{year}{2016}),
  \bibinfo{pages}{1157--1163}.
\newblock
\urldef\tempurl%
\url{https://doi.org/10.1177/0278364915620033}
\showDOI{\tempurl}


\bibitem[\protect\citeauthoryear{Engel, Koltun, and Cremers}{Engel
  et~al\mbox{.}}{2018}]%
        {engel2018direct}
\bibfield{author}{\bibinfo{person}{Jakob Engel}, \bibinfo{person}{Vladlen
  Koltun}, {and} \bibinfo{person}{Daniel Cremers}.}
  \bibinfo{year}{2018}\natexlab{}.
\newblock \showarticletitle{Direct Sparse Odometry}.
\newblock \bibinfo{journal}{\emph{IEEE Transactions on Pattern Analysis and
  Machine Intelligence (TPAMI)}} \bibinfo{volume}{40}, \bibinfo{number}{3}
  (\bibinfo{year}{2018}), \bibinfo{pages}{611--625}.
\newblock
\urldef\tempurl%
\url{https://doi.org/10.1109/TPAMI.2017.2658577}
\showDOI{\tempurl}


\bibitem[\protect\citeauthoryear{Engel, Sch{\"o}ps, and Cremers}{Engel
  et~al\mbox{.}}{2014}]%
        {engel2014lsdslam}
\bibfield{author}{\bibinfo{person}{Jakob Engel}, \bibinfo{person}{Thomas
  Sch{\"o}ps}, {and} \bibinfo{person}{Daniel Cremers}.}
  \bibinfo{year}{2014}\natexlab{}.
\newblock \showarticletitle{{LSD--SLAM}: Large-Scale Direct Monocular {SLAM}}.
  In \bibinfo{booktitle}{\emph{Proceedings of European Conference on Computer
  Vision (ECCV)}}. \bibinfo{pages}{834--849}.
\newblock
\urldef\tempurl%
\url{https://doi.org/10.1007/978-3-319-10605-2_54}
\showDOI{\tempurl}


\bibitem[\protect\citeauthoryear{Geiger, Lenz, and Urtasun}{Geiger
  et~al\mbox{.}}{2012}]%
        {geiger2012arewe}
\bibfield{author}{\bibinfo{person}{Andreas Geiger}, \bibinfo{person}{Philip
  Lenz}, {and} \bibinfo{person}{Raquel Urtasun}.}
  \bibinfo{year}{2012}\natexlab{}.
\newblock \showarticletitle{Are we ready for Autonomous Driving? The {KITTI}
  Vision Benchmark Suite}. In \bibinfo{booktitle}{\emph{Proceedings of IEEE
  Conference on Computer Vision and Pattern Recognition (CVPR)}}.
  \bibinfo{pages}{3354--3361}.
\newblock
\urldef\tempurl%
\url{https://doi.org/10.1109/CVPR.2012.6248074}
\showDOI{\tempurl}


\bibitem[\protect\citeauthoryear{Grisetti, K{\"u}mmerle, Stachniss, and
  Burgard}{Grisetti et~al\mbox{.}}{2010}]%
        {grisetti2010tutorial}
\bibfield{author}{\bibinfo{person}{Giorgio Grisetti}, \bibinfo{person}{Rainer
  K{\"u}mmerle}, \bibinfo{person}{Cyrill Stachniss}, {and}
  \bibinfo{person}{Wolfram Burgard}.} \bibinfo{year}{2010}\natexlab{}.
\newblock \showarticletitle{A Tutorial on Graph-Based SLAM}.
\newblock \bibinfo{journal}{\emph{IEEE Transactions on Intelligent
  Transportation Systems Magazine}} \bibinfo{volume}{2}, \bibinfo{number}{4}
  (\bibinfo{year}{2010}), \bibinfo{pages}{31--43}.
\newblock
\urldef\tempurl%
\url{https://doi.org/10.1109/MITS.2010.939925}
\showDOI{\tempurl}


\bibitem[\protect\citeauthoryear{K{\"u}mmerle, Grisetti, Strasdat, Konolige,
  and Burgard}{K{\"u}mmerle et~al\mbox{.}}{2011}]%
        {kummerle2011g2o}
\bibfield{author}{\bibinfo{person}{Rainer K{\"u}mmerle},
  \bibinfo{person}{Giorgio Grisetti}, \bibinfo{person}{Hauke Strasdat},
  \bibinfo{person}{Kurt Konolige}, {and} \bibinfo{person}{Wolfram Burgard}.}
  \bibinfo{year}{2011}\natexlab{}.
\newblock \showarticletitle{g2o: A general framework for graph optimization}.
  In \bibinfo{booktitle}{\emph{Proceedings of IEEE International Conference on
  Robotics and Automation (ICRA)}}. \bibinfo{pages}{3607--3613}.
\newblock
\urldef\tempurl%
\url{https://doi.org/10.1109/ICRA.2011.5979949}
\showDOI{\tempurl}


\bibitem[\protect\citeauthoryear{Moulon, Monasse, Marlet, et~al\mbox{.}}{Moulon
  et~al\mbox{.}}{2019}]%
        {moulon2019openmvg}
\bibfield{author}{\bibinfo{person}{Pierre Moulon}, \bibinfo{person}{Pascal
  Monasse}, \bibinfo{person}{Renaud Marlet}, {et~al\mbox{.}}}
  \bibinfo{year}{2019}\natexlab{}.
\newblock \bibinfo{title}{OpenMVG: An Open Multiple View Geometry library}.
\newblock \bibinfo{howpublished}{\url{https://github.com/openMVG/openMVG}}.
\newblock


\bibitem[\protect\citeauthoryear{Mu{\~{n}}oz{-}Salinas and {Medina
  Carnicer}}{Mu{\~{n}}oz{-}Salinas and {Medina Carnicer}}{2019}]%
        {rafael2019ucoslam}
\bibfield{author}{\bibinfo{person}{Rafael Mu{\~{n}}oz{-}Salinas} {and}
  \bibinfo{person}{Rafael {Medina Carnicer}}.} \bibinfo{year}{2019}\natexlab{}.
\newblock \bibinfo{title}{UcoSLAM: Simultaneous Localization and Mapping by
  Fusion of KeyPoints and Squared Planar Markers}.
\newblock
\newblock
\urldef\tempurl%
\url{https://doi.org/10.13140/RG.2.2.31751.65440}
\showDOI{\tempurl}
\showeprint{1902.03729}


\bibitem[\protect\citeauthoryear{Mur-Artal, Montiel, and Tard\'os}{Mur-Artal
  et~al\mbox{.}}{2015}]%
        {murartal2015orbslam}
\bibfield{author}{\bibinfo{person}{Ra\'ul Mur-Artal}, \bibinfo{person}{J.~M.~M.
  Montiel}, {and} \bibinfo{person}{Juan~D. Tard\'os}.}
  \bibinfo{year}{2015}\natexlab{}.
\newblock \showarticletitle{{ORB--SLAM}: a Versatile and Accurate Monocular
  {SLAM} System}.
\newblock \bibinfo{journal}{\emph{IEEE Transactions on Robotics}}
  \bibinfo{volume}{31}, \bibinfo{number}{5} (\bibinfo{year}{2015}),
  \bibinfo{pages}{1147--1163}.
\newblock
\urldef\tempurl%
\url{https://doi.org/10.1109/TRO.2015.2463671}
\showDOI{\tempurl}


\bibitem[\protect\citeauthoryear{Mur-Artal and Tard\'os}{Mur-Artal and
  Tard\'os}{2017}]%
        {murartal2017orbslam2}
\bibfield{author}{\bibinfo{person}{Ra\'ul Mur-Artal} {and}
  \bibinfo{person}{Juan~D. Tard\'os}.} \bibinfo{year}{2017}\natexlab{}.
\newblock \showarticletitle{{ORB--SLAM2}: an Open-Source {SLAM} System for
  Monocular, Stereo and {RGB-D} Cameras}.
\newblock \bibinfo{journal}{\emph{IEEE Transactions on Robotics}}
  \bibinfo{volume}{33}, \bibinfo{number}{5} (\bibinfo{year}{2017}),
  \bibinfo{pages}{1255--1262}.
\newblock
\urldef\tempurl%
\url{https://doi.org/10.1109/TRO.2017.2705103}
\showDOI{\tempurl}


\bibitem[\protect\citeauthoryear{Rosten and Drummond}{Rosten and
  Drummond}{2006}]%
        {rosten2006machine}
\bibfield{author}{\bibinfo{person}{Edward Rosten} {and} \bibinfo{person}{Tom
  Drummond}.} \bibinfo{year}{2006}\natexlab{}.
\newblock \showarticletitle{Machine Learning for High-Speed Corner Detection}.
  In \bibinfo{booktitle}{\emph{Proceedings of European Conference on Computer
  Vision (ECCV)}}. \bibinfo{pages}{430--443}.
\newblock
\urldef\tempurl%
\url{https://doi.org/10.1007/11744023_34}
\showDOI{\tempurl}


\bibitem[\protect\citeauthoryear{Rosten, Porter, and Drummond}{Rosten
  et~al\mbox{.}}{2010}]%
        {rosten2010faster}
\bibfield{author}{\bibinfo{person}{Edward Rosten}, \bibinfo{person}{Reid
  Porter}, {and} \bibinfo{person}{Tom Drummond}.}
  \bibinfo{year}{2010}\natexlab{}.
\newblock \showarticletitle{Faster and Better: A Machine Learning Approach to
  Corner Detection}.
\newblock \bibinfo{journal}{\emph{IEEE Transactions on Pattern Analysis and
  Machine Intelligence (TPAMI)}} \bibinfo{volume}{32}, \bibinfo{number}{1}
  (\bibinfo{year}{2010}), \bibinfo{pages}{105--119}.
\newblock
\urldef\tempurl%
\url{https://doi.org/10.1109/TPAMI.2008.275}
\showDOI{\tempurl}


\bibitem[\protect\citeauthoryear{Rublee, Rabaud, Konolige, and Bradski}{Rublee
  et~al\mbox{.}}{2011}]%
        {rublee2011orb}
\bibfield{author}{\bibinfo{person}{Ethan Rublee}, \bibinfo{person}{Vincent
  Rabaud}, \bibinfo{person}{Kurt Konolige}, {and} \bibinfo{person}{Gary
  Bradski}.} \bibinfo{year}{2011}\natexlab{}.
\newblock \showarticletitle{{ORB}: An efficient alternative to {SIFT} or
  {SURF}}. In \bibinfo{booktitle}{\emph{Proceedings of IEEE International
  Conference on Computer Vision (ICCV)}}. \bibinfo{pages}{2564--2571}.
\newblock
\urldef\tempurl%
\url{https://doi.org/10.1109/ICCV.2011.6126544}
\showDOI{\tempurl}


\bibitem[\protect\citeauthoryear{Schlegel, Colosi, and Grisetti}{Schlegel
  et~al\mbox{.}}{2018}]%
        {schlegel2018proslam}
\bibfield{author}{\bibinfo{person}{Dominik Schlegel}, \bibinfo{person}{Mirco
  Colosi}, {and} \bibinfo{person}{Giorgio Grisetti}.}
  \bibinfo{year}{2018}\natexlab{}.
\newblock \showarticletitle{{ProSLAM: Graph SLAM from a Programmer's
  Perspective}}. In \bibinfo{booktitle}{\emph{Proceedings of IEEE International
  Conference on Robotics and Automation (ICRA)}}. \bibinfo{pages}{1--9}.
\newblock
\urldef\tempurl%
\url{https://doi.org/10.1109/ICRA.2018.8461180}
\showDOI{\tempurl}


\bibitem[\protect\citeauthoryear{Sch\"{o}nberger and Frahm}{Sch\"{o}nberger and
  Frahm}{2016}]%
        {schoenberger2016sfm}
\bibfield{author}{\bibinfo{person}{Johannes~Lutz Sch\"{o}nberger} {and}
  \bibinfo{person}{Jan-Michael Frahm}.} \bibinfo{year}{2016}\natexlab{}.
\newblock \showarticletitle{Structure-from-Motion Revisited}. In
  \bibinfo{booktitle}{\emph{Proceedings of IEEE Conference on Computer Vision
  and Pattern Recognition (CVPR)}}. \bibinfo{pages}{4104--4113}.
\newblock
\urldef\tempurl%
\url{https://doi.org/10.1109/CVPR.2016.445}
\showDOI{\tempurl}


\bibitem[\protect\citeauthoryear{Sturm, Engelhard, Endres, Burgard, and
  Cremers}{Sturm et~al\mbox{.}}{2012}]%
        {sturm2012benchmark}
\bibfield{author}{\bibinfo{person}{Jrgen Sturm}, \bibinfo{person}{Nikolas
  Engelhard}, \bibinfo{person}{Felix Endres}, \bibinfo{person}{Wolfram
  Burgard}, {and} \bibinfo{person}{Daniel Cremers}.}
  \bibinfo{year}{2012}\natexlab{}.
\newblock \showarticletitle{A Benchmark for the Evaluation of RGB-D SLAM
  Systems}. In \bibinfo{booktitle}{\emph{Proceedings of IEEE/RSJ International
  Conference on Intelligent Robots and Systems (IROS)}}.
  \bibinfo{pages}{573--580}.
\newblock
\urldef\tempurl%
\url{https://doi.org/10.1109/IROS.2012.6385773}
\showDOI{\tempurl}


\bibitem[\protect\citeauthoryear{Sweeney, Hollerer, and Turk}{Sweeney
  et~al\mbox{.}}{2015}]%
        {sweeney2015theia}
\bibfield{author}{\bibinfo{person}{Christopher Sweeney},
  \bibinfo{person}{Tobias Hollerer}, {and} \bibinfo{person}{Matthew Turk}.}
  \bibinfo{year}{2015}\natexlab{}.
\newblock \showarticletitle{Theia: A Fast and Scalable Structure-from-Motion
  Library}. In \bibinfo{booktitle}{\emph{Proceedings of the 23rd ACM
  International Conference on Multimedia}}. \bibinfo{pages}{693--696}.
\newblock
\urldef\tempurl%
\url{https://doi.org/10.1145/2733373.2807405}
\showDOI{\tempurl}


\bibitem[\protect\citeauthoryear{Umeyama}{Umeyama}{1991}]%
        {umemiya1991least}
\bibfield{author}{\bibinfo{person}{Shinji Umeyama}.}
  \bibinfo{year}{1991}\natexlab{}.
\newblock \showarticletitle{Least-Squares Estimation of Transformation
  Parameters Between Two Point Patterns}.
\newblock \bibinfo{journal}{\emph{IEEE Transactions on Pattern Analysis and
  Machine Intelligence (TPAMI)}} \bibinfo{volume}{13}, \bibinfo{number}{4}
  (\bibinfo{year}{1991}), \bibinfo{pages}{376--380}.
\newblock
\urldef\tempurl%
\url{https://doi.org/10.1109/34.88573}
\showDOI{\tempurl}


\end{thebibliography}

\end{document}